\documentclass[pdflatex,sn-mathphys, 10pt]{sn-jnl}
\usepackage{setspace}
\usepackage{caption}
\usepackage{subcaption}

\begin{document}

\title{Neuradicon: operational representation learning of neuroimaging reports}

\author*[1]{\fnm{Henry} \sur{Watkins}}\email{h.watkins@ucl.ac.uk}
\author[1]{\fnm{Robert} \sur{Gray}}
\author[1]{\fnm{Adam} \sur{Julius}}
\author[2]{\fnm{Yee-Haur} \sur{Mah}}
\author[2]{\fnm{James} \sur{Teo}}
\author[2]{\fnm{Walter H.L.} \sur{Pinaya}}
\author[2]{\fnm{Paul} \sur{Wright}}
\author[1]{\fnm{Ashwani} \sur{Jha}}
\author[1]{\fnm{Holger} \sur{Engleitner}}
\author[2]{\fnm{Jorge} \sur{Cardoso}}
\author[2]{\fnm{Sebastien} \sur{Ourselin}}
\author[3]{\fnm{Geraint} \sur{Rees}}
\author[1]{\fnm{Rolf} \sur{Jaeger}}
\author*[1]{\fnm{Parashkev} \sur{Nachev}}\email{p.nackev@ucl.ac.uk}

\affil*[1]{\orgdiv{Queen Square Institute of Neurology}, \orgname{University College London}, \orgaddress{\city{London},\country{United Kingdom}}}
\affil[2]{\orgdiv{School of Biomedical Engineering \& Imaging Sciences}, \orgname{King's College London}, \orgaddress{\city{London}, \country{United Kingdom}}}
\affil[3]{\orgname{University College London}, \orgaddress{\city{London},\country{United Kingdom}}}
\maketitle
\newpage
\abstract{
\textbf{Background and Objective}
\newline
Radiological reports typically summarize the content and interpretation of imaging studies in unstructured form that precludes quantitative analysis. This limits the monitoring of radiological services to throughput undifferentiated by content, impeding specific, targeted operational optimization. Here we present Neuradicon, a natural language processing (NLP) framework for quantitative analysis of neuroradiological reports.
\newline
\textbf{Methods}
\newline
Our framework is a hybrid of rule-based and machine-learning models to represent neurological reports in succinct, quantitative form optimally suited to operational guidance. These include probabilistic models for text classification and tagging tasks, alongside auto-encoders for learning latent representations and statistical mapping of the latent space. 
\newline
\textbf{Results}
\newline
We demonstrate the application of Neuradicon to operational phenotyping of a corpus of 336,569 reports, and report excellent generalizability across time and two independent healthcare institutions. In particular, we report pathology classification metrics with f1-scores of 0.96 on prospective data, and semantic means of interrogating the phenotypes surfaced via latent space representations. 
\newline
\textbf{Conclusions}
\newline
Neuradicon allows the segmentation, analysis, classification, representation and interrogation of neuroradiological reports structure and content. It offers a blueprint for the extraction of rich, quantitative, actionable signals from unstructured text data in an operational context.
}

\doublespacing

\section*{Abstract}
\textbf{Background and Objective}
\newline
Radiological reports typically summarize the content and interpretation of imaging studies in unstructured form that precludes quantitative analysis. This limits the monitoring of radiological services to throughput undifferentiated by content, impeding specific, targeted operational optimization. Here we present Neuradicon, a natural language processing framework for quantitative analysis of neuroradiological reports.
\newline
\textbf{Methods}
\newline
Our framework is a hybrid of rule-based and machine-learning models to represent neurological reports in succinct, quantitative form optimally suited to operational guidance. These include probabilistic models for text classification and tagging tasks, alongside auto-encoders for learning latent representations and statistical mapping of the latent space. 
\newline
\textbf{Results}
\newline
We demonstrate the application of Neuradicon to operational phenotyping of a corpus of 336,569 reports, and report excellent generalizability across time and two independent healthcare institutions. In particular, we report pathology classification metrics with f1-scores of 0.96 on prospective data, and semantic means of interrogating the phenotypes surfaced via latent space representations. 
\newline
\textbf{Conclusions}
\newline
Neuradicon allows the segmentation, analysis, classification, representation, and interrogation of neuroradiological reports structure and content. It offers a blueprint for the extraction of rich, quantitative, actionable signals from unstructured text data in an operational context.
\newline
\textbf{Keywords:}
Natural Language Processing, Neurology, Neuroradiology, Artificial Intelligence


\section{Introduction}
\label{S:intro}

Neuroradiological reports capture a crucial aspect of the management of patients with neurological disorders, providing radiologist-defined interpretations of imaging as part of the broader electronic health record (EHR) \cite{Esteva2019}. Although reporting practices increasingly include structured elements, the complexity of the task compels free prose as the dominant form of communication. This complicates the task of quantitative analysis of radiological content---as opposed to volume---on which the evidence-guided operational optimization of radiological services inevitably depends. A report processing system that can reliably extract operationally-relevant information is fundamental to capturing service workload with adequate fidelity.  

The ideal tool for improving such workloads is the natural language processing (NLP) \cite{Jurafsky:2009:SLP:1214993}. This subfield of machine learning applies various models to free text to extract meaningful or relevant information. In particular, most models work by tokenizing the text into word-like token sub-sequences. With these we can perform several tasks such as text classification, token classification (closely related to named-entity-recognition), text segmentation, along with a set of other specific tasks to extract useful information from a text.  

Recent developments in the field of NLP  have opened the door to extracting quantitative representations of complex clinical texts, enabling excellent performance in real-world settings \cite{wu2020deep, bobba2023natural} . Indeed, the widespread use of language models outside of medicine has inspired some to suggest a revolution in medical NLP \cite{lastrucci2024revolutionizing}. The task nonetheless remains challenging, especially where---as in neuroradiology---both vocabulary and logical grammar may be unusual, and the minimal acceptable standard of fidelity is high. The components of the task generally consist of tokenisation, named entity recognition (NER), and negation detection, along with aligning entities to a structured ontology \cite{Pons2016} or other domain-specific frameworks such as the system developed by Jorg et al. \cite{jorg2023efficient}.  The key difficulty in training and validating machine learning models here is the paucity of domain-specific training data, and the complexity of aggregating data at scale given privacy and security constraints\cite{Sheikhalishahi2019}. Moreover, manual labelling of clinical reports is time-consuming and requires specialist domain knowledge of low availability and high cost. The sensitivity of clinical data is complicated by the difficulty of reliable anonymisation in a setting where proper names have a multiplicity of uses besides reference to an individual (e.g. eponymous syndromes).

In the field of neuroradiology, the information of operational interest divides across pathological appearances, their anatomical locations, instrumental factors of relevance to interpretation, and such diagnostic conclusions as the appearances justify. In particular, knowing \textit{what} has been observed \textit{where}, enables stratification of the imaged population by machine-generated latent representational phenotypes, database queries of individual features of operational concern, such as contrast utilization, and modelling of the distribution of patient radiological appearances. Rich featurisation of reports can both guide actions relevant to service delivery and provide a mechanism for interrogating the contents of the associated image within multi-modal models of text and imaging, amplifying the operational value of the system. Here we present a pipeline crafted with these objectives in mind in the context of a service optimisation project at University College London Hospitals NHS Trust.

Numerical text representation methods such as word2vec \cite{Mikolov2013} and Glove \cite{pennington2014glove}, as well as term-frequency-inverse-document-frequency vectorisation (TF-IDF), have been used to classify texts using logistic regression or support vector machines (SVM). These map texts or tokens into a numerical vector space. These vector representations can be used by further models in tasks like text classification. Further works have introduced domain specific biomedical transformer models, like BioBert \cite{lee2020biobert} and BioGPT \cite{luo2022biogpt}. These models have been successfully applied to classification of lung radiology texts \cite{ong2023automated} and breast imaging reports \cite{tan2023natural}. Comparisons of different models in the field of radiology text interpretation has been performed by Liu et al. \cite{liu2023evaluating} and have identified both strengths and weaknesses in this domain. 

Aside from NER, another critical task in information extraction is negation detection. Here the aim is to classify a clinical concept present in the text as negated or not, an area of special importance in medicine where the absence of a relevant feature may be just as important as its presence, and in equal need of explicit statement. One rules-based method to achieve this in the field of medical text is Negbio \cite{peng2018negbio}. Negbio uses grammatical patterns across the dependency parse tree to evaluate whether a concept is negated or not. The universal dependency tree was designed to provide a tree description of the grammatical relationships in a sentence and used by downstream language processing tasks. All universal dependency information can be represented by a directed graph, the vertices of which are labelled with the word and part-of-speech. Work by Sykes et al. \cite{Sykes2021} has found that while machine learning approaches are effective, rules-based methods like Negbio perform equally well. 

The grammatical dependency parse tree can also be used for relation extraction, whereby we relate two entities tagged by an NER model using their grammatical relationship. RelEx, introduced by Fundel et al. \cite{Fundel2007}, showed how this method could extract pairs of entities in a biomedical text with a particular semantic relationship. Deep learning methods for relation extraction have also been explored, with Li et al. \cite{Li2019} using a neural network to model the shortest dependency path in clinical text. A recent review by Zhao et al. \cite{zhao2024comprehensive} documents the application of LLMs to this task, including the application to biomedical text, where they have now become the norm.

In neuroradiology, rules-based systems such as EDIE-R \cite{Alex2019} are able to extract entities and relations, and have been used to phenotype patients using radiological reports \cite{Wheater2019}. An extension to neural-network methods found approximately equal performance on a neuroradiological dataset \cite{Gorinski2019}. While these results are promising for neuroradiology, we aim to leverage the progress with language models and domain-specific pre-training to create a comprehensive NLP pipeline. Some works, such as Chappidi et al. \cite{chappidi2024defining} have focussed on using LLMs for extraction tasks. However, the ubiquity of language models for NLP tasks in recent years has been shown to be inadequate in tasks with small or imbalanced datasets \cite{yang2023transformer}, hence the use of such a hybrid pipeline. This pipeline needs not only to extract named entities and clinical concepts but also to use context-aware numerical representations to classify texts, to recognise negation, to segment reports into relevant sections, and to discover whether a report is compared with previous imaging.

Here we describe `Neuradicon', a framework for deriving rich, quantitative representations of neuroradiological free text reports with the purpose of facilitating the content-aware operational optimization of neuroradiological services. Neuradicon performs a sequence of tasks, each with an associated model, combining rule-based, deep learning, and traditional machine-learning algorithms. While deep learning methods can lead to great improvements in performance across many tasks, conventional methods are sufficient for many components the pipeline and reduce the minimal computational footprint, an important consideration in resource-challenged healthcare environments. The complete list of tasks is as follows:

\begin{enumerate}
\item Report classification: classifying reports free from pathological appearances, or rendered potential elliptic by comparison with previous imaging. This stage acts as a filter, to identify interpretable reports.
\item Section classification: assigning sentences to separate sections of the report.
\item Named entity recognition: finding tokens that are clinically relevant named entities.
\item Negation detection: classifying named entities as negated or not.
\item Relation extraction: discovering semantic relations between named entities in the text.
\item Pathological domain classification: classifying a report into one or more pathological domains.
\item Report clustering: where we find a low-dimensional representation of a report for unsupervised phenotyping and finding similar reports.
\end{enumerate}

Our paper proceeds as follows. Descriptions of our pipeline methods and data sets can be found in sections \ref{S:datacollect} to \ref{S:geospm}, with our phenotyping procedure defined in section \ref{S:phenotype}.
In section \ref{S:results} we describe the resulting performance of our pipeline on real cross-site clinical labelled data from both University College London Hospital (UCLH) and King's College Hospital (KCH). In section \ref{S:embed-results} we present the latent embedding space for appearances present in radiological reports; we describe the patterns in pathological appearances that are now clear when viewed in this unsupervised manner. Section \ref{S:spm-results} covers the application of spatial statistical inference to our neuroradiology space, where we identify key regions and explain their contents. In section \ref{S:pheno-results} we describe the results of the phenotyping. In section \ref{S:conclusions} we discuss the impact and application of this model in the context of practice and our results.
\begin{figure}[h]
\centering
\includegraphics[width=\linewidth]{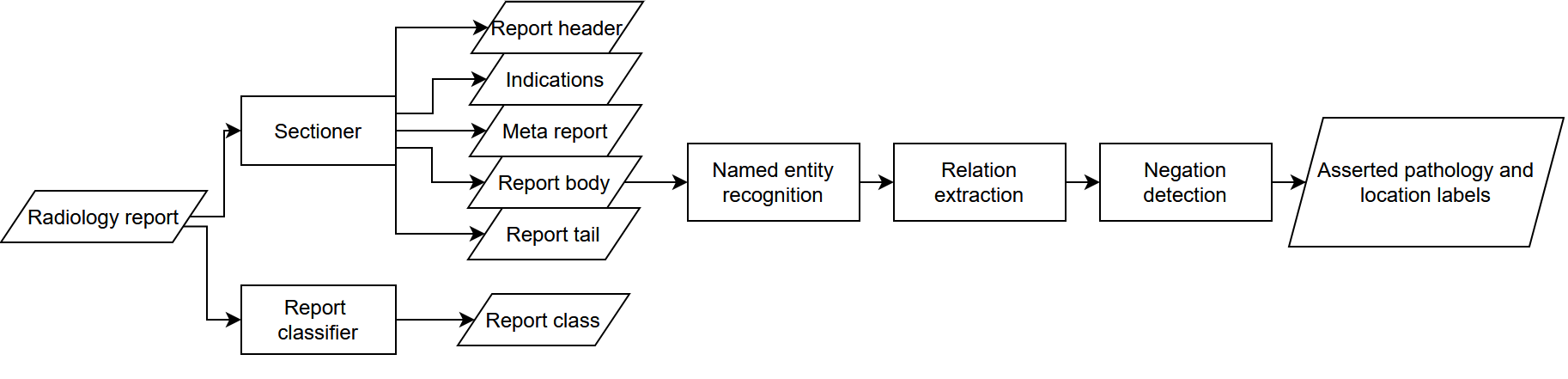}
\caption{A diagram of the NLP pipeline. The models, represented by rectangles, are a mixture of machine-learning and rules-based methods. The output data (romboids) from each model are passed to another in the pipeline.}\label{fig:pipeline}
\end{figure}

\section{Methods}
\label{S:methods}

\subsection{Data \& Labelling}
\label{S:datacollect}

The dataset for this work is a corpus of 336,569 anonymized radiological reports extracted from the PACS system of the National Hospital for Neurology and Neurosurgery (NHNN), London, UK, for the purpose of local service evaluation and optimization. This corpus represents scans for a total of 181,519 unique patients from August 1999 until February 2023. These patients have a mean age of 51.6 (standard deviation=17.1) and are $44.3\%$ male and $55.7\%$ female. Samples of reports were extracted from this dataset for hand labelling, while the remaining reports were used for unsupervised training of a custom language model. The data for this project was extracted in its raw format, as unstructured free text. Empty reports were retained as we also needed the model to deal effectively with empty or missing reports. An example report can be found in Appendix \ref{app:report} for a brain MRI for a patient with hydrocephalus.

For each of the seven tasks, we have produced an accompanying labelled evaluation dataset drawn from a corpus of MRI neuroradiological reports. These are sourced from the UCLH National Hospital for Neurology and Neurosurgery (NHNN). In addition to UCLH data, we also have external evaluation data from King's College Hospital (KCH). This data was used to verify the cross-site performance. When data is sourced from a single site, the resulting training and test data can be idiosyncratic, limiting domain transfer. It is necessary to include external data from another site to ensure the models have not overfitted to the style and structure of UCLH reporting. The datasets used for this project are summarized in Table \ref{tab:data}. For full details concerning the labelling tool, bias mitigation and task-specific dataset construction, see Appendix \ref{app:data}.

\begin{table}
\begin{center}
\caption{Evaluation data sets. The eight tasks tested are or 5 different types. Each task corresponds to a different sub-model of the pipeline}\label{tab:data}
\begin{tabular}{@{}lll@{}}
\toprule
Task & Type  & N reports\\
\midrule
Report classification  &  multi-class classification &  400\\
Section segmentation & token classification   &  165 \\
Negation detection & entity-wise binary classification & 287\\
Relation extraction & pair-entity-wise binary classification &  120\\
Pathological domain classification (UCLH) & multi-label classification &  572 \\
Prospective domain classification (UCLH) & multi-label classification &  170 \\
Pathological domain classification (KCH) & multi-label classification &   526 \\
Prospective domain classification (KCH) & multi-label classification &  537 \\
\botrule
\end{tabular}
\end{center}
\end{table}

\subsection{Custom Language Model}
\label{S:word2vec}

The foundation for the deep learning tasks in the pipeline is numerical representations of tokens. These n-dimensional feature vectors are then fed into downstream tasks such as classification and named entity recognition. We created a custom language model by fine tuning the BioBERT language model \cite{lee2020biobert} on a dataset of 200,000 radiological reports from our corpus. The 200,000 reports extracted at random from the full dataset as a fine-tuning set. These were further split into train-evaluate split (80\%/20\%) for the training routine.
We started with the BioBERT base model and continued training using our data to create a model that is tuned to the domain-specific data we are using. To train our model, we used the huggingface transformers library \cite{jain2022hugging}. While the advances in generative language models has been the subject of significant research, such as Biogpt \cite{luo2022biogpt} or Med-PaLM \cite{tu2023towards}, the purpose of this language model is for downstream representation in discriminative probabilistic models. With no need for text generation, the BERT architecture is well suited to our tasks. The reason for training the BioBERT model further on an extract of our radiological texts is because domain-specific training has been shown to improve overall in-domain performance \cite{gu2021domain}. Further details describing the model can be found in Appendix \ref{app:llm}.
`
\subsection{Report Classification}
Our first model in the pipeline implements a text classification model. These models take the numerical representation of a text sequence provided by the underlying language model, and train a multi-class classifier neural network model on top of the aggregated token vectors. A key feature in clinical decision-making and filtering reports for operational studies is the normality of a report. In the context of neuroradiological reporting, a normal report is devoid of any asserted abnormal neurological appearances. This information can be used as a radiologist-defined label for the imaging associated with this report and identifying it as a healthy brain. With such a label, one could train AI clinical image models on the large datasets of MRI and CT imaging that exist for clinical scans. Another valuable source of information in a radiological report is whether it is a comparative. Ascertaining whether a radiological report is being compared to previous imaging is valuable for the standard radiological workflow. However, often it is only within the text itself that there is an indication that the current image is being compared to previous imaging. For example, many reports in the corpus are considered comparative because they contain phrases like ``Comparison is made with the previous scan performed 6 April 2016."

The full classification task aims to classify texts into five exclusive classes:
\begin{enumerate}
\item ABNORMAL: A report where the radiologist has noted pathological appearances, e.g. ``There are multiple supertentorial lesions".
\item NORMAL FOR AGE: A report that has noteworthy features, but not outside what is expected considering the patient's age, e.g. ``There is some age-related volume loss"
\item NORMAL: A report where commentary denotes a normal scan, e.g. ``normal appearances", or wherein all pathological appearances are negated, e.g. ``There is no sign of a tumour".
\item COMPARATIVE: A report compared to previous imaging. This class is exclusive, because it implies there is another report with additional commentary. The nature of radiological reporting means radiologists will omit commentary on appearances that are mentioned in previous reports. As the report is not complete, we cannot conclude the report is NORMAL or ABNORMAL.
\item MISSING: A report that is empty, or a placeholder. e.g. ``No report generated for this image". The reality of in-situ radiological reporting means we must account for missing information and mistakes. 
\end{enumerate}

These classes were so chosen because they inform how reporting information is used in downstream operational use. The distinction between NORMAL and NORMAL FOR AGE, while seemingly academic, reflects the actions taken by clinicians in practice. Likewise, the distinction between COMPARATIVE and ABNORMAL is necessary because the potential omission of information in a comparative report must be considered. The downside of these classes is the classification problem is harder than a simple binary healthy/unhealthy classification, but it is necessary when considering real-world data. The MISSING class is included to deal with missing and incomplete reports. Real-world medical data inevitably includes corrupted or missing data, and this part of the pipeline also filters out missing data through classification of the incomplete reports. Further details describing the model can be found in Appendix \ref{app:cls}

\subsection{Section Classification}
For the section classification task, the aim is to classify sentences of the report into the following report sections: 

\begin{enumerate}
\item header - the scan protocol and machine-generated information such as the scan number and date.
\item indications - detail of relevant patient medical history and indications.
\item metareport - commentary on the report itself, such as the quality of the scan, or whether it is compared to previous imaging.
\item body - the main content of the report detailing the findings of the scan.
\item tail - radiological reports are conventionally signed by the author.
\end{enumerate}

\begin{figure}
    \centering
    \includegraphics[width=\linewidth]{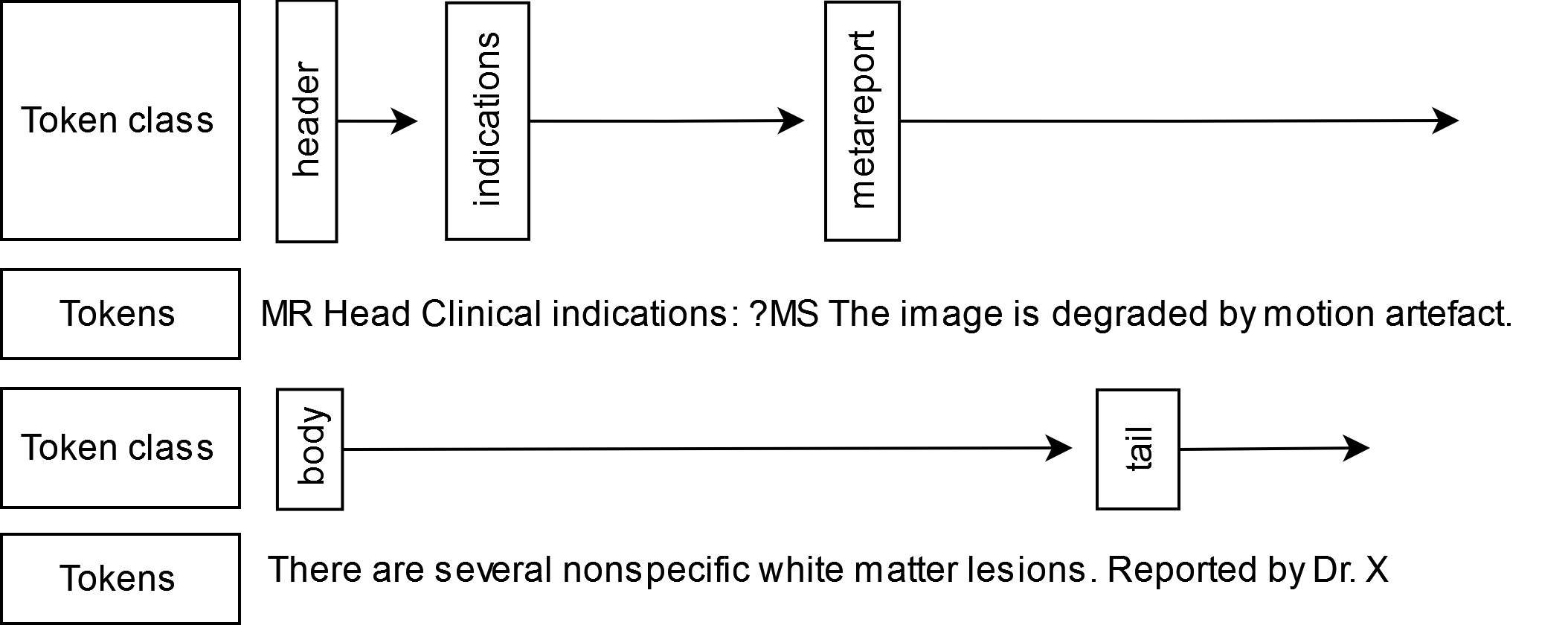}
    \caption{The schema for the report section classification and segmentation model. The first token of the report section is classified into one of 5 classes. This acts as the section 'anchor' token, and all subsequent tokens belong to the same section. Once all tokens have been classified to a section, we can segment the report into meaningful mutually-exclusive sections.}
    \label{fig:section}
\end{figure}

It is essential to segment the report into these sections to aid information extraction and ensure that any extracted information can be placed into the right context in which it is used. This task is a token classification problem. In our architecture, the report is segmented by classifying certain tokens as the `section start' tokens, with all subsequent tokens belonging to that class, see Figure. \ref{fig:section} for an illustration. 
The model classifies each token by taking the activations from the custom language model and learning a shallow classifier on top. Further details describing the model can be found in Appendix \ref{app:cls2}

\subsection{Named Entity Recognition}
\label{S:ner}
The task of named entity recognition in the field of neurology is to recognise clinical concepts present within the report. In practice, this task implements a token classification model, whereby each token vector produced by the underlying language model has a multi-class classifier trained on top of it. The tokens not only is labelled by particular classes, but uses the BIO scheme whereby each token is labelled as the beginning of an entity (B), within and entity (I) or outside an entity (O). For the purposes of neuroradiology, we choose the following broad categories:

\begin{enumerate}
\item pathology - a clinical concept that refers to a particular neurological condition
\item location - a particular anatomical location. 
\item descriptor - a clinical concept that is used to describe the appearances of the patient's scan in the context of abnormality.
\end{enumerate}

Within each of these classes, we have chosen a set of subdomains that are important in the context of representing neurological reporting. These classes are summarised in Table \ref{tab:domains}.
\begin{table}
\begin{center}
\caption{Neuroradiological entity classes. Each entity mentioned in the reports can be classed into three broad classes. The entities and their respective classes are identified by an NER model}\label{tab:domains}
\begin{tabular}{@{}lll@{}}
\toprule
Pathology & Descriptor  & Locations\\
\midrule
haemorrhagic  &  cyst &  arteries\\
ischaemic & damage  &  brain stem \\
vascular & diffusion & diencephalon\\
other cerebrovascular & signal change  &  ear, nose \& throat \\
treatment \& surgery & enhancement &  eye \\
inflammatory \& autoimmune & flow-related &  ganglia \\
congenital \& developmental & interval \& change &   grey matter \\
csf disorders & mass \& effect &  limbic system \\
musculoskeletal & morphology &  meninges \\
neoplastic \& paraneoplastic & collection &  nerves \\
infectious & necrosis &  neurosecretory system \\
neurodegenerative \& dementia &  &  skull \\
metabolic, nutritional \& toxic &  &  spine \\
endocrine &  &  telencephalon \\
ophthalmological &  &  veins \\
traumatic & & ventricles \\
& & white matter\\
& & other location \\
\botrule
\end{tabular}
\end{center}
\end{table}
These domains reflect broad classes of disorders encountered in neurology, and they are each constituted by a set of conditions. For example, the `pathology-treatment' class is constituted by terms such as `post surgical change', `shunt', `surgery', `biopsy', `chemotherapy', `surgical defect', `resection', `cranioplasty', among many others that have been extracted from the reports themselves and assigned to pathological domains.

The architecture for the NER task is also token classification model, using the same implementation and training routine as the section classification model. However, for NER a token aggregation strategy is required to fuse tokens into entities based on model predictions. For our model we use the 'first'  token strategy, where the class of the first token is used for the entity class when there is ambiguity. An example of a report tagged by the NER task is shown in Figure. \ref{fig:ner} Further details describing the model can be found in Appendix \ref{app:ner}
\begin{figure}[h!]
\centering\includegraphics[width=\linewidth]{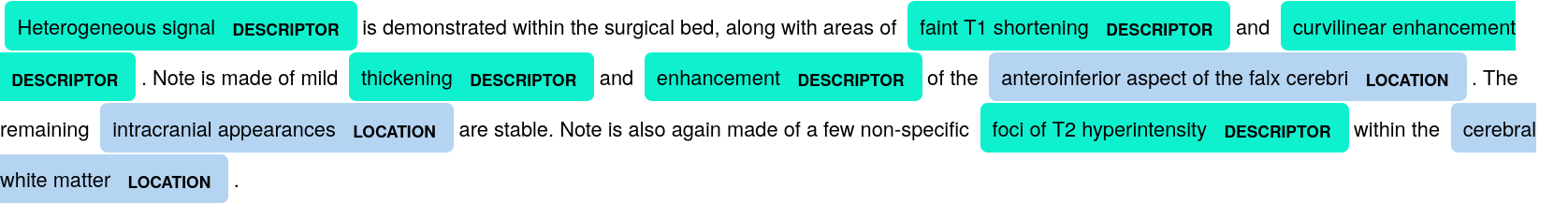}
\caption{This example of a radiological report shows words tagged with named entities. This image shows the broad classes of DESCRIPTOR, LOCATION and PATHOLOGY. The NER model classifies tokens according to the BIOS format. }
\label{fig:ner}
\end{figure}
\subsection{Negation Detection}
Negation detection is performed on each entity produced by the NER task. For this task, we implement the Negbio rule-based method \cite{peng2018negbio}. The Negbio method uses patterns on universal grammatical dependencies \cite{Jurafsky:2009:SLP:1214993} to ascertain whether a clinical finding is negated. The algorithm for this task uses a modified set of Semgrex \cite{Chambers2007} patterns to extract tokens that satisfy particular grammatical patterns. Semgrex patterns are strings that encode patterns on a universal dependency parse tree. During development, the full list of patterns used in Negbio was found to be unnecessary for adequate performance on the stereotyped structure of neuroradiological reports. We used a reduced set specific to the field of neurology. For this task, we used the grammatical dependency parse produced by the Spacy parser as input and implemented the patterns using the Spacy dependency matcher. Since the negation classification task is a rule-based method, no training is necessary.  

Several of the operators used below are not in the Semgrex standard set, for more information about the implementation of these patterns, see www.spacy.io. Note this pattern set is simpler than that presented in Negbio due to the more stereotyped structure of neuroradiological reporting.

The patterns are:
\begin{verbatim}
1. {ner:/PATHOLOGY|DESCRIPTOR/}>
        {dep:/neg|det|case/&lemma:/no|not|none|without/}
2. {ner:/PATHOLOGY|DESCRIPTOR/}$--
        {dep:/neg|det|case/&lemma:/no|not|none|without/}
3. {ner:/PATHOLOGY|DESCRIPTOR/}<<
        {ner:/PATHOLOGY|DESCRIPTOR/}$--
            {dep:/neg|det|case/&lemma:/no|not|none|without/}
4. {ner:/PATHOLOGY|DESCRIPTOR/}<{}$--
        {ner:/PATHOLOGY|DESCRIPTOR/}>
            {dep:/neg|det|case/&lemma:/no|not|none|without/}
5. {ner:/PATHOLOGY|DESCRIPTOR/&dep:/conj|acl|nmod|amod|dobj/}<<
        {ner:/PATHOLOGY|DESCRIPTOR/}>
            {dep:/neg|det|case/&lemma:/no|not|none|without/}
6. {ner:/PATHOLOGY|DESCRIPTOR/};*
        {lemma:/evidence|finding|focus|sign|feature/}>
            {dep:/neg|det|case/&lemma:/no|not|none|without/}
\end{verbatim}
The patterns in this task focus only on entities of the pathological and descriptive classes (see Table \ref{tab:domains} for their contents).  This is because radiological commentary consists primarily of the presence or absence of neurological appearances. As such, anatomical entities, though present in reports, are described as the location of appearances rather than the subject of reporting itself.

\subsection{Relation Extraction}

The relation extraction task aims to match named entities of the `pathology' and `descriptor' classes to their corresponding anatomical locations, if present. This approach is narrower in scope than the general task of relation extraction found in the literature \cite{Jurafsky:2009:SLP:1214993} and is tailored to the needs of neuroradiology. We base our method on the RelEx system \cite{Fundel2007}, but in our work, we use relation extraction for one particular relation: entity lies within a location. Like the negation detection algorithm, the relation extraction algorithm uses the grammatical dependency parse of a sentence to extract the relations. This pipe searches for patterns on the grammatical dependency parse tree; pairs of entities are linked by a `within' relation if a `location' class named entity and a `pathology' or `descriptor' named entity satisfy one of these grammatical patterns. Since the relation extraction task is a rule-based method, no training is necessary. Like the negation task, the patterns for the relation extraction task are represented using Semgrex \cite{Chambers2007} patterns. These patterns prescribe a path along the dependency parse tree starting at the PATHOLOGY or DESCRIPTOR entity, and traversing the path until a LOCATION entity is reached. Several of the operators used below are not in the Semgrex standard set, for more information about the implementation of these patterns, see www.spacy.io. Further details describing the model can be found in Appendix \ref{app:rel}

The patterns are:
\begin{verbatim}
1. {ner:/PATHOLOGY|DESCRIPTOR}>>{ner:/LOCATION/}
2. {ner:/PATHOLOGY|DESCRIPTOR/}\$++{ner:/LOCATION/}
\end{verbatim}
\subsection{Report-Level Pathological Domains}
\label{S:domains}
The primary goal of this complete pipeline is to identify pathological labels for reports. This means using the outputs from previous tasks to assign multiple non-exclusive binary labels to each report. Pathological domains are assigned to reports according to whether they have at least one asserted pathology entity of that class. Thus each report has potentially several of the following pathological domain labels. The identification of the report-level binary labels is thus criterial, and depends solely on the output of the NER and negation models. Since we have also identified `descriptor' and `location' entity types in the named entity recognition section \ref{S:ner}, we can also extract labels for anatomical locations and radiological appearances.

\subsection{Latent Space Embedding}
\label{S:embedding}
The latent space representation of radiological reporting in neurology was obtained by training an auto-encoder model \cite{hinton1993autoencoders} to map a binary representation of reports (from section \ref{S:domains}) to a 2-dimensional vector space. Autoencoders are an unsupervised model and so the latent space is not constructed but learned from the underlying structure of the data itself. The diversity of appearances in reporting is captured by the training procedure. By learning the best model that reconstructs the original data, the autoencoder learns a low-dimensional latent space representation of the data structure. While a simpler linear dimensionality reduction method such as PCA can be used to obtain 2D representations of high-dimensional data, it is not able to capture the complex non-linear interactions available to a deep autoencoder. By using a non-linear dimensionality reduction method we can capture the complex non-linear patterns present in language data. This makes them more effective for extracting meaningful results from our datasets.

The procedure has four stages:
\begin{enumerate}
\item Aggregating semantically similar pathological entity features into super-groups,
\item Creating a high-dimensional binary vector representation of a report
\item Combining these with the super-group and pathological domain labels
\item Training an deep auto-encoder to learn a 2-dimensional real-valued latent representation
\end{enumerate}

Each report is passed through the comprehensive NLP pipeline. This pipeline tags each report with the pathological entities present, and determines whether they are asserted or not. We extract all of the asserted pathological entities and convert these to a binary vector representation, whereby each entry in a binary vector records the presence of that pathological term in the corresponding report. This is done by a n-gram text vectorizer from the scikit-learn library. 

Although each report is now represented in a numerical format,  we have so far ignored that fact that multiple terms may be semantically similar or synonymous. We aggregate terms together by performing k-means clustering of the word vectors learned in section \ref{S:word2vec}. Each feature identified in the vectorization process has a point in the word vector space. We then cluster these points into 100 groups. This means for each report we have two representations, a `low-level' binary representation where every unique asserted pathological entity has an entry, and a smaller `high-level' binary representation where the constituent asserted pathology terms are `binned' into larger semantic groups; each element in this binary vector represents the presence of one of these higher level groups. Finally, we concatenate these two binary vectors together, so that each report has both a `low-level' and `high-level' representation.

These aggregated binary vectors are fed into an autoencoder that is trained to project the high-dimensional binary data down to a 2-dimensional semantic embedding. The choice of a 2 dimensional embedding was primarily for purposes of inspection and visualisation of the latent space. A latent space of 3, 4 or more dimensions would improve reconstruction performance during training, but we couldn't intuit the distribution nor plot the resulting latent vectors. Further, we find a low-dimensional bottleneck is necessary to force the model to learn a semantically meaningful representation and aid interpretability of the model outputs. The embedding model is illustrated in Figure \ref{fig:embed}. Full details of the autoencoder model structure and training can be found in Appendix \ref{app:ae}.
\begin{figure}
    \centering
    \includegraphics[width=\linewidth]{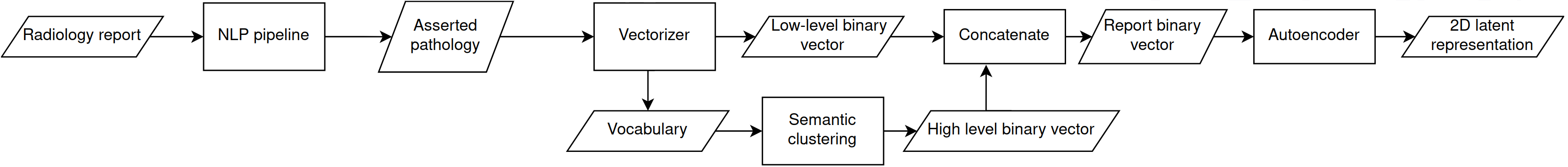}
    \caption{A diagram of the report embedding model and procedure. Text reports are featurized by the NLP pipeline. entities of interest are extracted and these turned into binary vectors ore being embedded into a 2d latent space by an auto-encoder. }\label{fig:embed}
\end{figure}

\subsection{Identifying Phenotypes}
\label{S:phenotype}

Once a latent space representation is obtained, we would like to find the characteristic features of a particular group or cluster of reports in the latent space. Because the auto-encoder forces the binary report representations into a small 2d real-valued bottleneck, we can interpret the euclidean distance in this latent space as a similarity metric that incorporates the distribution of patterns of features across the whole set of reports. Thus reports that cluster together in the latent space have similar appearances, and describe patients with similar conditions (and co-occurrences of conditions). This enables us to identify particular phenotypes of patients in an unsupervised manner, one that has been learned from the data itself, rather than according to a prescriptive criterion.

We identify the features that identify the phentotype by calculating the in-group vs out-group phi-correlation coefficient. Also known as the Matthews correlation coefficient, this measure produces a correlation coefficient for each binary feature in the input, and allows us to discover the `characteristic' features of particular clusters and subregions of the latent space. Given an cluster subset of reports selected from the latent space - ideally a set that show a clear clustering - we construct the contingency table for each binary feature in the initial binary featurization. 

For example, if we consider a binary pathological feature, say `tumour', each report has a binary vector with a row indicating whether they contain this feature or not. With a sample of 100 reports, we assign 13 to a group or cluster; with these we could create a contingency table for each feature. Our example table would measure the intersection of in vs. out cluster and with vs. without tumour. This contingency table gives us the phi-coefficient for each feature in the input. The highest-phi features are those most strongly correlated with the cluster, and we can identify the cluster by these features - we use these features to define our cluster phenotype. See section \ref{S:pheno-results} for an example.

\subsection{Spatial Inference}
\label{S:geospm}

With a learned latent space of pathological appearances, we can use spatial inference to identify regions that are identified with certain variables. For this task we make use of geoSPM \cite{engleitner2022geospm}, a method of applying statistical parametric mapping to spatial data. It is a means of performing topological inference by extending the statistical parametric mapping approach \cite{friston1994statistical} of the SPM software to a space of 2D points.

Our corpus of reports are mapped to a point in a 2d latent space; each representing a collection of (potential) pathological appearances. GeoSPM runs topological analysis on this space to derive marginalised spatial maps, disentangling appearances and variables of interest from confounding factors. The primary value of this analysis is to identify regions where particular variables are significant. It also allows us to find the conjunctions of multiple variables, representing regions where multiple factors are jointly significant. 

For our analysis, geoSPM software was run with a resolution of 256x199, with a smoothing level of 0.4. The areas of significance for individual spatially distributed variables are identified by areas where the local regression coefficient value reaches a significance level of $\alpha=0.05$ in a two-tailed t-test.

In order to ensure valid topological conclusions in the spatial inference process, we implemented several methodological safeguards. First, we used GeoSPM to construct statistically thresholded t-maps in our learned 2D latent space, allowing us to identify subregions significantly modulated by variables of interest such as age or treatment. We then carefully controlled for potential confounders by introducing these variables into multiple regression models, thus isolating the effects in question while mitigating spurious associations. Moreover, we confirmed the semantic coherence of the latent space by demonstrating alignment with established pathological domains (e.g., ischemic, hemorrhagic, neoplastic) and validating these clusters against known clinical patterns (e.g., ICD-10 codes, contrast usages). Finally, plausibility checks against well-established clinical knowledge further reinforced that our spatial inferences reflect genuine relationships rather than modeling artifacts. The results of this analysis cna be found in section \ref{S:geospm}.

\section{Results}
\label{S:results}
\subsection{Summary}
In this section we present a comprehensive analysis of the NLP pipeline's capabilities and insights into neurological patterns. The pipeline exhibited good performance across all major tasks, with particularly strong outcomes in report classification and section classification. When projected onto the latent space, our analysis uncovered intricate relationships between different neurological conditions. Notably, ischaemic and hemorrhagic domains showed significant overlap, reflecting shared pathological processes. Within the hemorrhagic region, cavernomas formed a distinct, dense cluster, while aneurysms emerged as a specific subset. This in turn reveals relationships between diseases. For example, we see complex interactions between treatment and neoplastic cases, with clear clustering patterns emerging in specific spatial regions. Age-related patterns showed strong correlations with both ischaemic conditions and treatment domains, with distinct age-related structures appearing within specific treatment regions. 
The cross-site validation demonstrated robust model performance between different hospital settings, with strong pathological domain classification at both UCLH and KCH sites. The model's ability to maintain high performance across different clinical settings underscores its potential for widespread clinical application.
The latent representation successfully identified distinct imaging phenotypes, revealing patterns of co-occurring conditions such as small vessel disease with infarction, demyelination with volume loss, and cerebral atrophy with small vessel disease. These phenotypes enable a compressed yet rich representation of neuroradiological activity, facilitating the identification of distinct subpopulations where variation in outcomes needs specific attention.

\subsection{Performance Metrics}
The data sets described in \ref{S:datacollect} allow us to evaluate the performance of each of the respective information extraction tasks. The performance for these tasks on held-out data, both prospective and cross-site, is measured in terms of micro precision-recall-f1 metrics presented in Table \ref{tab:metrics}. 

\begin{table}[h]
\begin{center}
\caption{Performance metrics for pipeline models. Each of the models in the pipeline has a task and corresponding labelled dataset. The performance for each task is evaluated in terms of the precision, recall and f1-score}\label{tab:metrics}
\begin{tabular}{@{}lccc@{}}
\toprule
Task & precision  & recall & f1-score\\
\midrule
Report classification  & 0.96  &  0.96 & 0.96 \\
Section classification & 0.93 & 0.92 & 0.93 \\
Negation detection & 0.97  & 0.90 & 0.93 \\
Relation extraction & 0.82 & 0.77 & 0.79\\
Pathological domain classification (UCLH) & 0.96 & 0.93 & 0.94 \\
Pathological domain classification (KCH) & 0.92 & 0.83 & 0.87 \\
Prospective domain classification (UCLH) & 0.97 & 0.96 & 0.96 \\
Prospective domain classification (KCH) & 0.95 & 0.97 & 0.96 \\
\bottomrule
\end{tabular}
\end{center}
\end{table}
Note the ultimate task here is pathological domain classification, where excellent fidelity is maintained both across time and site, despite inevitable variations in reporting styles and digital documentation systems. 

Upon inspection of the trained classifier models we identified that despite good cross-site classification performance for the domain classifier models, class-specific over-fitting was qualitatively present for particular classes.

\subsection{Representation learning}
\label{S:embed-results}
The entire corpus of UCLH neuroradiological reports was embedded into a 2-dimensional latent space with a deep auto-encoder of asserted pathological terms as described in \ref{S:embedding}. This comprehensive representation can be viewed as a surveyable `universe' of neuroradiology, organized into constellations of characteristic patterns of pathological appearances as described and interpreted by radiologists. Annotating the representation with individual variables---both internal and external---provides an indication of the drivers of the underlying structure, and applying topological inference allows us to demarcate regions of the manifold significantly associated with characteristics of operational interest.   

\begin{figure}[h]
\centering
\includegraphics[scale=0.35]{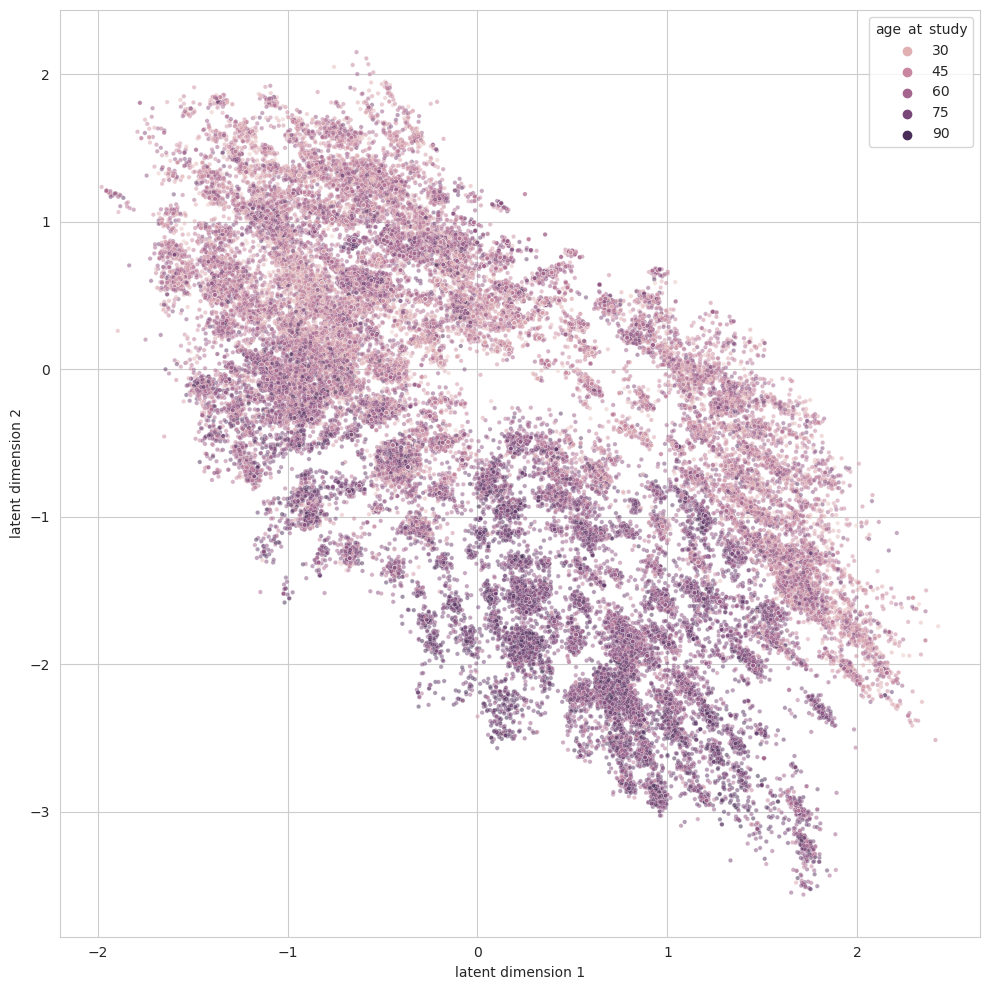}
\caption{The 2D latent representation of asserted pathological terms, labelled by patient age. Reports are projected onto a 2d latent space using an auto-encoder, based on their pathology textual content. Note variation in age-related patterns across the space}\label{fig:cluster}
\end{figure}

\begin{figure}
\centering
\includegraphics[width=\textwidth]{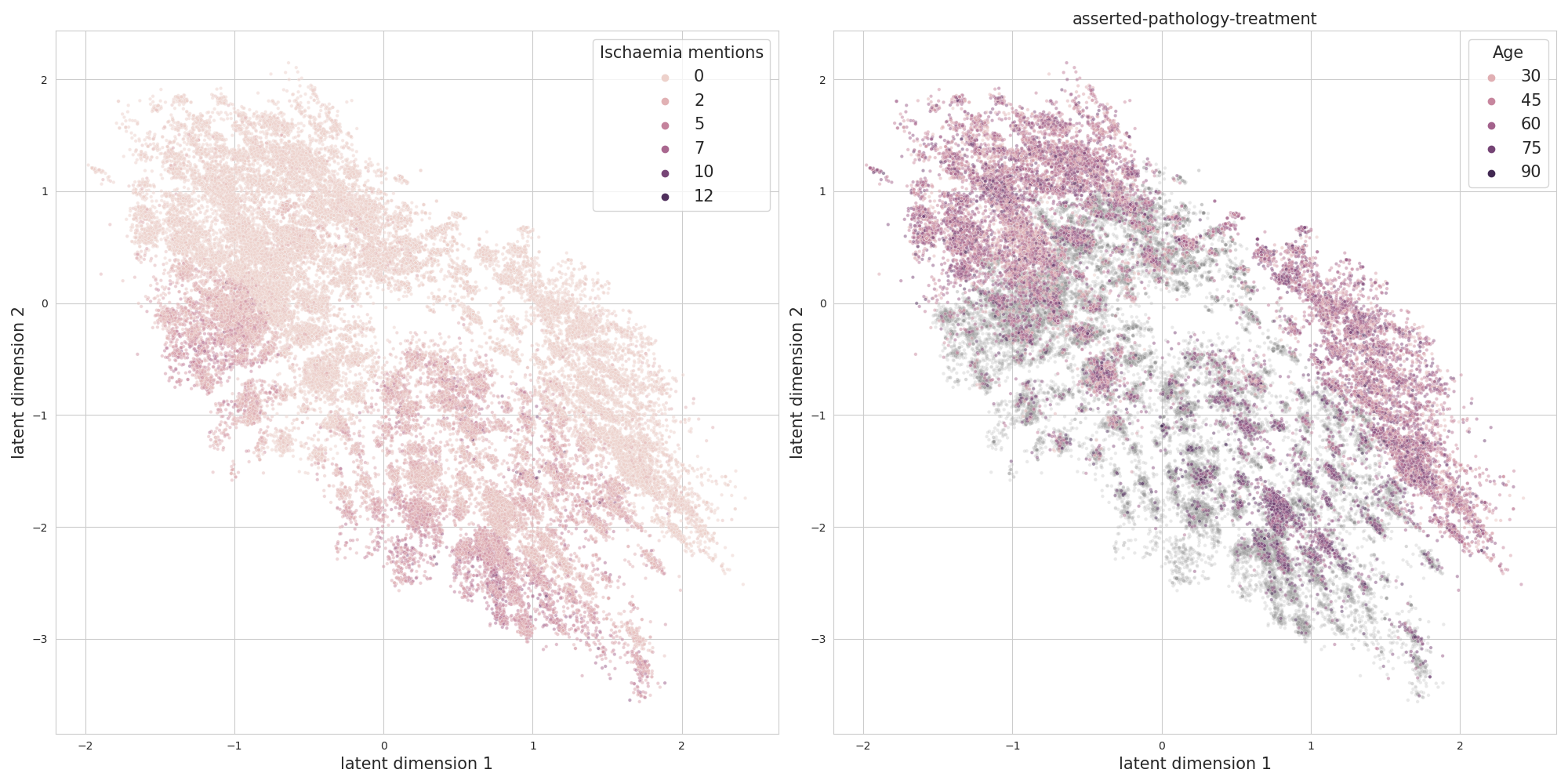}
\caption{The latent representation of radiological reports labelled by the number of asserted ischaemia-related terms (L). Reports embedded in a 2d space coloured by age show the relationship between age and the treatment domain (with non-treatment reports greyed out) (R). }
\label{fig:ischaemic-treatment}
\end{figure}

Age, shown in Figure. \ref{fig:cluster}, a variable external to the model, reveals variation in age-related patterns of pathological appearances across the latent space. 
Visual inspection of the latent space reveals patterns of organisation coherent with the ontological structure of neuroimaging appearances. Disentangled areas of the representation are identifiable for the most common domains: ischaemic, haemorrhagic, neurodegenerative, inflammatory, and neoplastic (with and without treatment). Figure. \ref{fig:ischaemic-treatment} shows a clear subregion of ischaemia-dense reports concentrated in the lower half of the space. Pathological domains that are rarer or more heterogeneous in their description are less well demarcated. Relations between domains are inspectable through joint labelling, e.g. age and treatment as illustrated in Figure. \ref{fig:ischaemic-treatment}, where colouring the reports of the treatment class by age shows clearly demarcated regions, some with a dominant age structure.

\begin{figure}
         \centering
         \includegraphics[width=\textwidth]{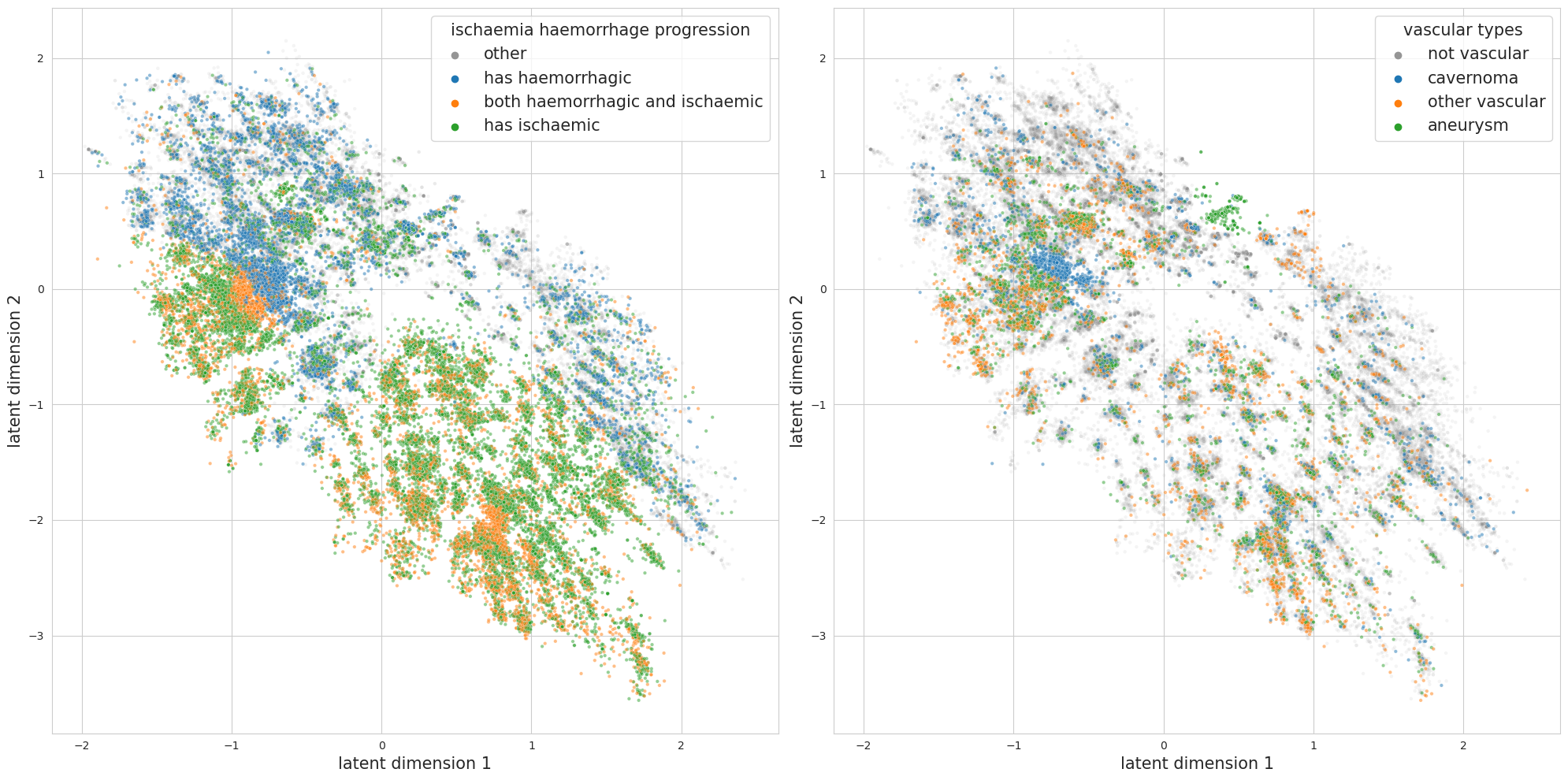}
         \caption{This plot latent-space reports coloured by pathology classes shows ischaemic and haemorrhagic domains intersect across the representational space in reflection of common pathological processes (L). Cavernomas represent a dense cluster within the wider haemorrhagic region, with aneurysms a subset of the haemorrhagic class. (R) }
         \label{fig:ischaemia-progression-vascular}
\end{figure}

\begin{figure}
         \centering
         \includegraphics[width=\textwidth]{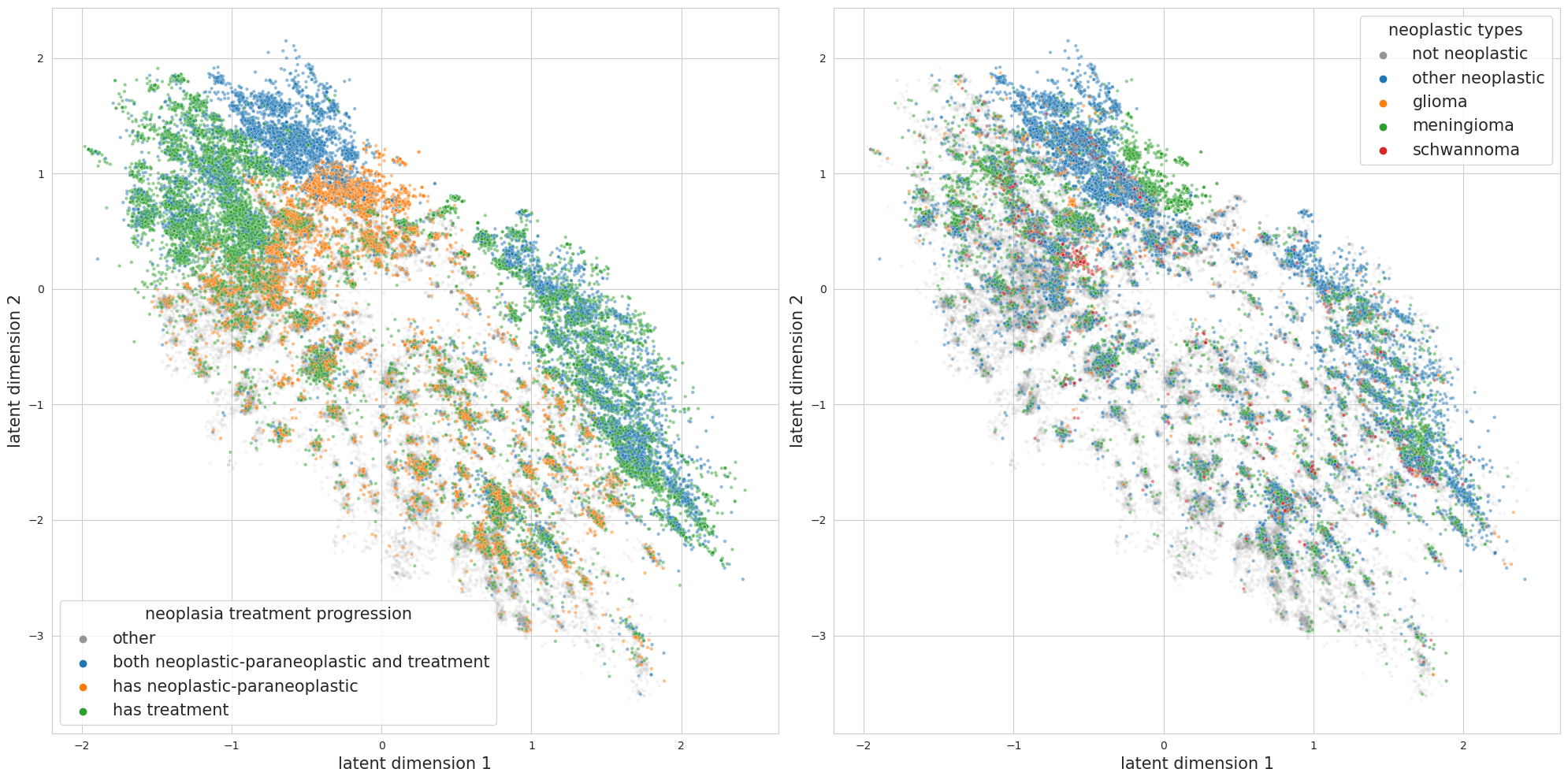}
         \caption{When projected into a 2d latent space, reports in the treatment class or the neoplastic class cluster in particular regions, including several areas of intersection between the two classes (L). Different tumour types distribute unevenly across the latent space. (R) }
         \label{fig:neoplasia-progression}
\end{figure}

\begin{figure}
\centering
\includegraphics[width=\textwidth]{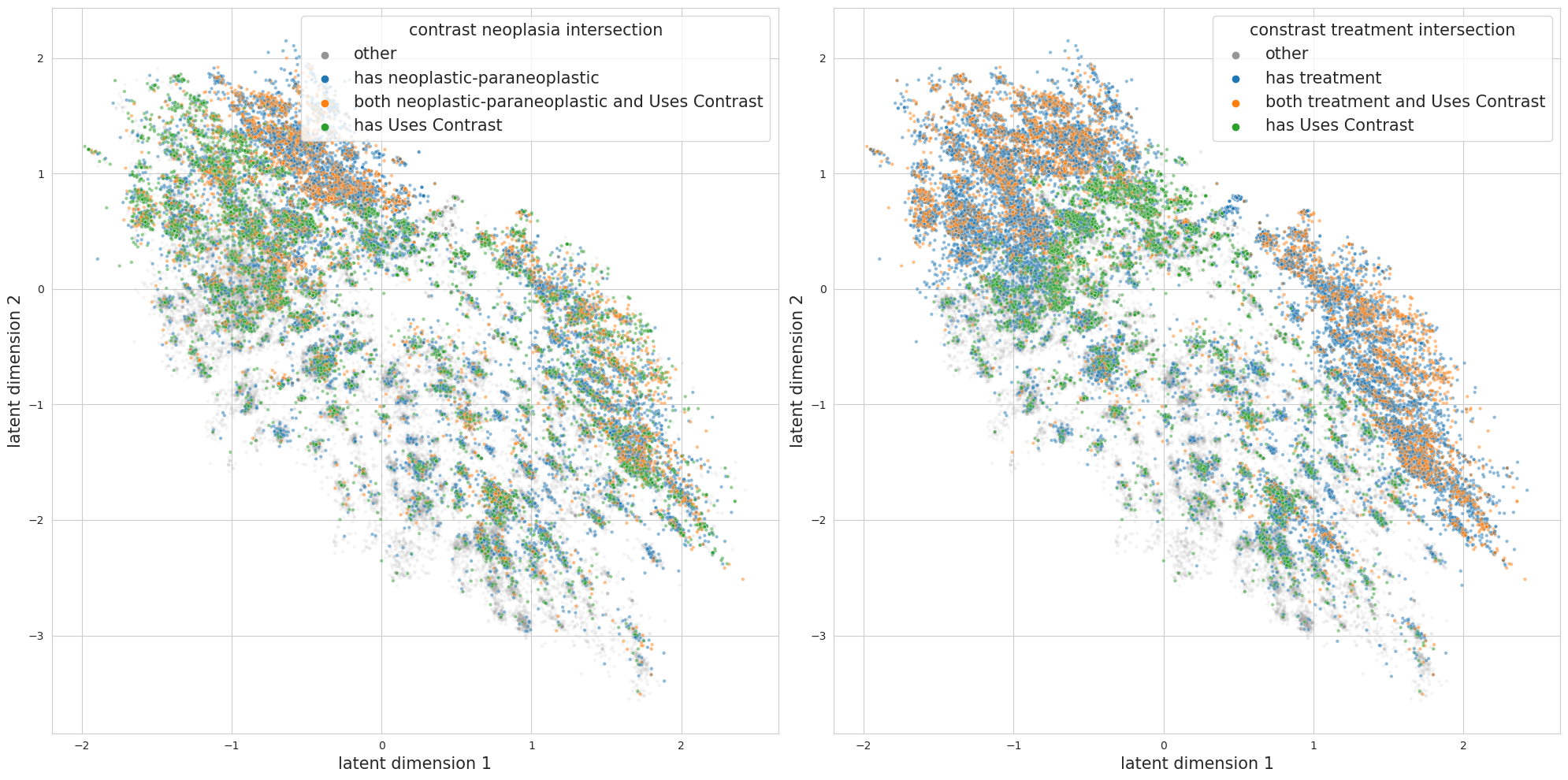}
\caption{The intersection between reports with contrast, belonging to the treatment domain (L). Contrast usage and particular pathological classes show significant intersection, reflecting clinical practice for particular conditions. (R)}
\label{fig:contrast-treatment}
\end{figure}

Ischaemic and treatment domains are the largest, the latter dominated by neoplasia. Ischaemic regions are differentiated by the temporality of clinical presentation, and the presence or absence of associated haemorrhagic changes. This is illustrated in Figure. \ref{fig:ischaemia-progression-vascular}. The heterogeneity of haemorrhagic changes, arising in the context of multiple distinct pathological processes, is evident in their wide distribution. Aneurysmal disease is distributed across the haemorrhagic and ischaemic regions, and with cavernomas represented as a clear sub-region of haemorrhagic disease.
The neoplastic domain commands a significant area of the latent space, with a broad overlap with treatment (Figure. \ref{fig:neoplasia-progression}). The territory is segregated by common tumour types, and diagnosis-specific interactions with treatment, e.g. well-defined areas for meningiomas and post surgical change.
Our model also allows for the stratification of reports by scan modality, use of contrast and ordering clinician, permitting the generation of operational insights around the use of specific treatment modalities and pathways in relation to the content of imaging reports. For example, Figure \ref{fig:contrast-treatment} shows the intersection of contrast usage and neoplastic or treatment classes. We can also identify the distribution of reports that describe a contrast-administered scan for patients with neoplastic or paraneoplastic conditions. We find a significant distribution of contrast usage across the space of reports in the treatment class.
\subsection{Spatial Inference}
\label{S:spm-results}
\begin{figure}
\centering
\includegraphics[width=\textwidth]{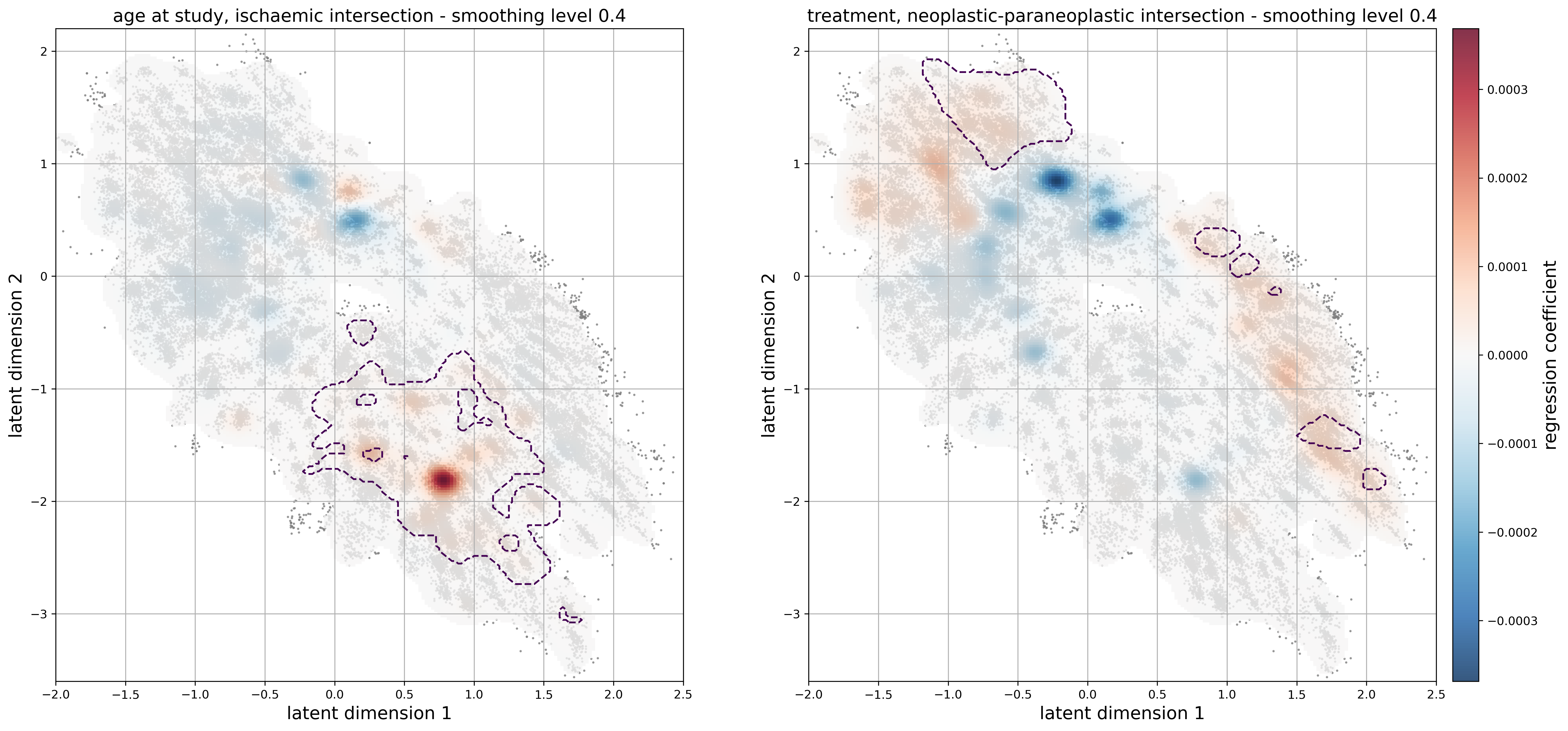}
\caption{The regression coefficient value for patient age, overlaid on the latent space. Superimposed is the region of statistically-significant conjunction between patient age and ischaemia overlaid as a contour (L). The regression coefficient value for the treatment pathology class, overlaid on the latent space. The region of statistically-significant conjunction between the areas of treatment and neoplasia is overlaid as a contour. (R)}
\label{fig:spm-age}
\end{figure}

We can use topological inference to formalise the task of identifying regions of interest. We use GeoSPM to derive statistical maps identifying sub-regions significantly associated with a given feature, controlled for the influence of confounding variables. The procedure is detailed in section \ref{S:geospm}. Figure \ref{fig:spm-age}  shows the regression coefficient maps for patient age and treatment domains. 

Spatial inference performed with geoSPM shows regions of high significance for particular spatially distributed variables.  These maps ground our intuition informed by the metadata-overlay figures. Figure \ref{fig:spm-age} identifies clear regions of statistically significant regression coefficient value for patient age and for treatment, indicating sub-regions where these features are particularly important. One can go further and identify topological relations between variables. GeoSPM can do this by mapping jointly modulated areas arising from the conjunction of two or more thresholded t-statistic maps.  These t-test maps show the regions of statistically signifiance with a level of $\alpha< 0.05$  .  Figure \ref{fig:spm-age} shows the conjunction of age and ischaemia, corresponding to the regions where age and more ischaemia coincide. We also identify clear islands jointly modulated by neoplasia and treatment.

\subsection{Imaging Report Phenotypes}
\label{S:pheno-results}
\begin{figure}[h]
\centering
\includegraphics[scale=0.35]{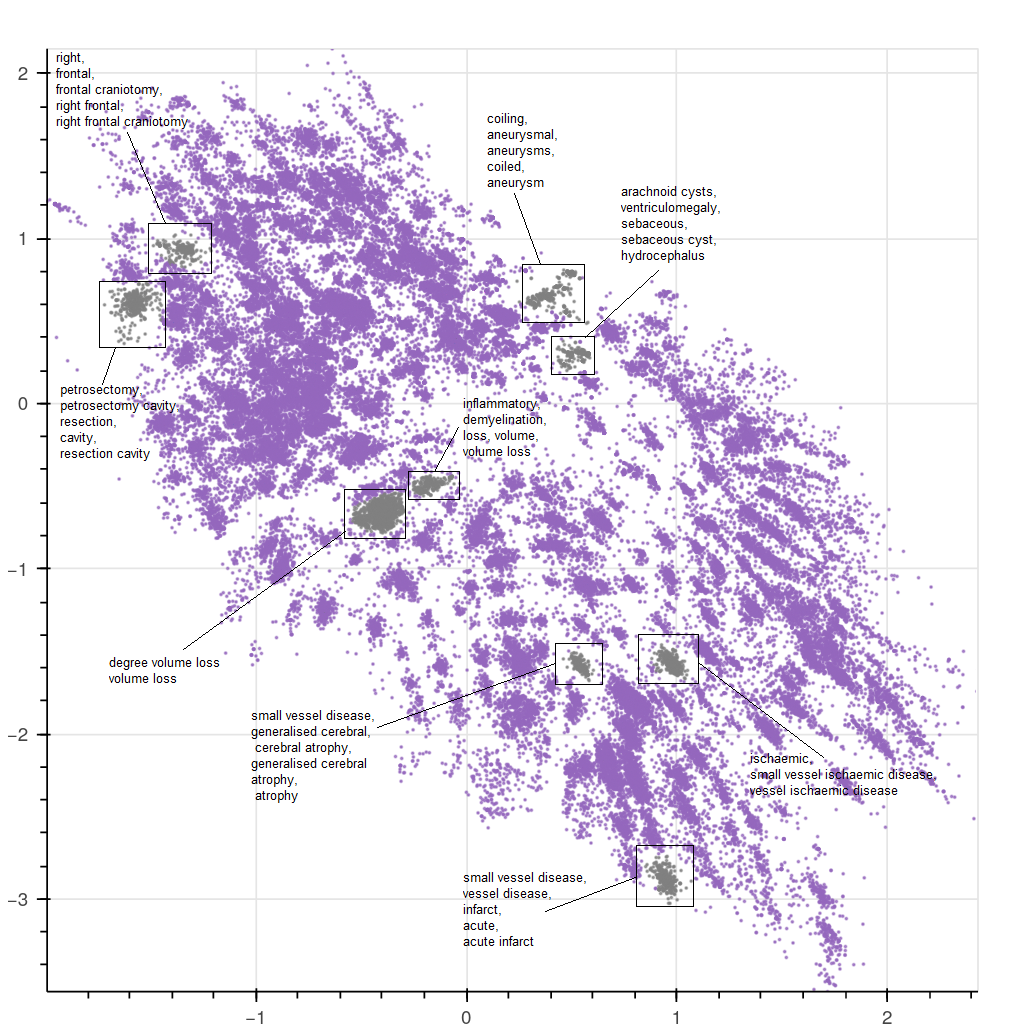}
\caption{When reports are embedded into a low-dimensional latent space, we can identify imaging phenotypes revealed by the structure of the latent representation. Plotted are several example clusters labelled by the characteristic pathological features uncovered by our pipeline. }\label{fig:phenotypes}
\end{figure}

The structure revealed in the latent representation provides a means of deriving a novel set of compressed 'report phenotypes' defined by characteristic patterns of covariance of individual input features (\ref{S:phenotype}). Figure \ref{fig:phenotypes} shows several example clusters and their characteristic features, as determined by the phenotyping model. For example, we find dense clusters dominated by craniotomy, coiled aneurysms, volume loss, and small vessel disease, but also clusters defined by co-occurring conditions, such small vessel disease and infarction, demyelination and volume loss, and cerebral atrophy and small vessel disease.

Imaging phenotypes enable the compressed representation of neuroradiological activity in terms of rich, yet succinct, descriptions of each report, facilitating the identification of distinct subpopulations where variation in an outcome of interest needs to be specifically detected to optimise service delivery. For example, mean reporting times may be found to be longer for a specific phenotype defined not by indication, or any other aspect external to the report, but the appearances themselves.   

\subsection{Clinical Coding}
\label{S:icd10}
\begin{figure}[h]
\centering
\includegraphics[width=\textwidth]{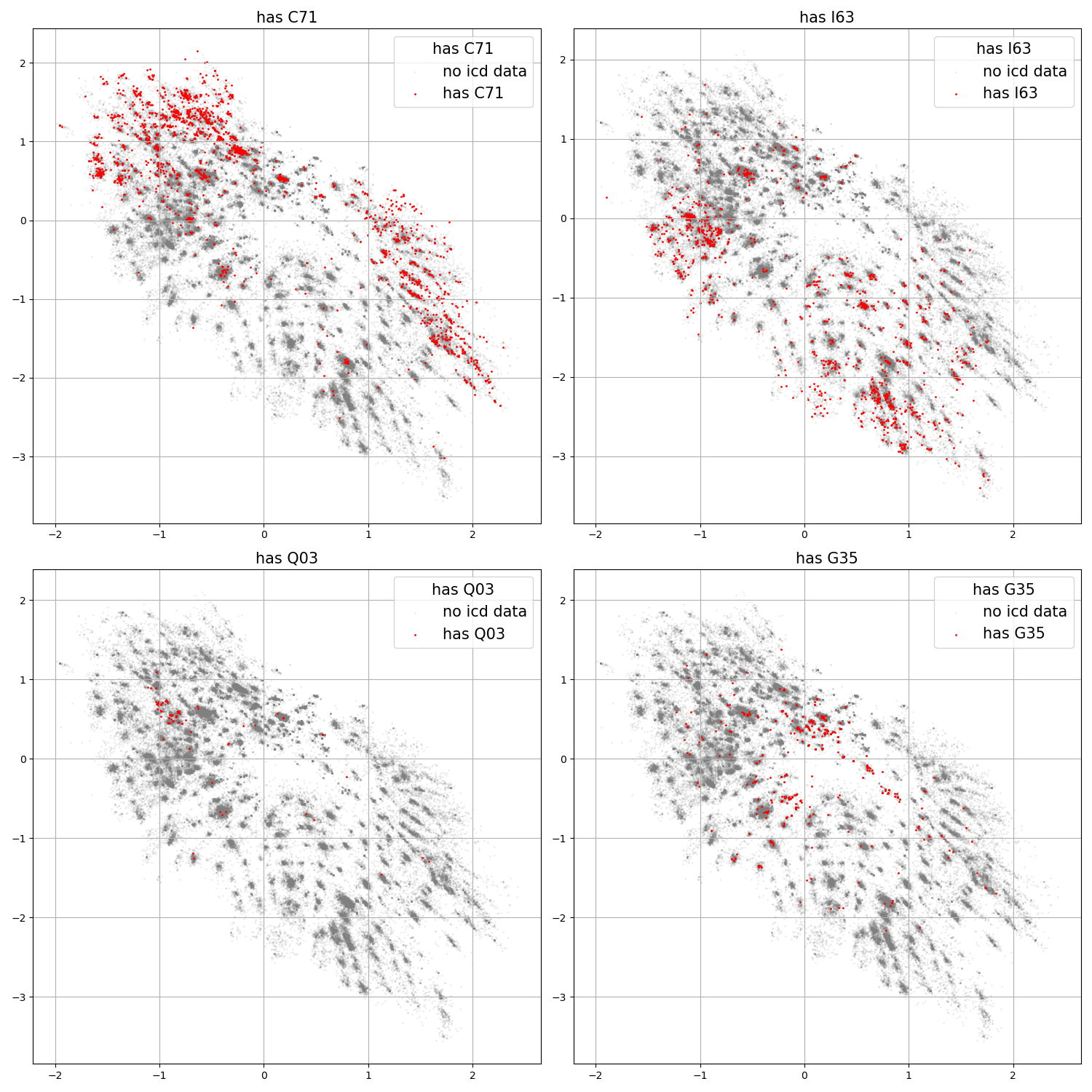}
\caption{The distribution of reports labelled with the icd10 code C71: Malignant neoplasm of brain (top left). The distribution of reports labelled with the icd10 code I63: Cerebral infarction (top right). The distribution of reports labelled with the icd10 code Q03:  Congenital hydrocephalus (bottom left). The distribution of reports labelled with the icd10 code G35: Multiple sclerosis (bottom right).}\label{fig:icd10_c71}
\end{figure}

Although neuroradiology typically provides only part of the evidence on which a diagnosis rests, a correspondence between the latent representation and ICD10 codes associated with the investigational episode should be expected \cite{world2004international}. Figure  \ref{fig:icd10_c71}, shows reports with codes belonging to four neurological conditions: C71 (Malignant neoplasm of brain), I63 (Cerebral infarction), Q03 (Congenital hydrocephalus) and G35 (Multiple sclerosis) available for a subset of episodes. These codes were chosen because they represent conditions with very different aetiologies, and roughly correspond with four of the most common pathological domains defined in section \ref{S:domains}. These are: pathology-ischaemic and I63; pathology-csf-disorders and Q03; pathology-inflammatory-autoimmune and G35; and pathology-neoplastic-paraneoplastic and C71.

For each of the ICD10 code plots, reports with each specific code occupy a subset of the total space. Comparison with these domains in sections \ref{S:embed-results} and section \ref{S:spm-results}  show qualitative intersections in the subspaces dominated by their corresponding ICD10 codes. It must be stressed that these codes were applied at the patient level, reflecting broader clinical management, whereas the latent space position is derived solely from the radiological report. While correlation between the two would be expected, the strongly localised distribution of codes into particular regions of the latent space indicates the connection between radiological description and patient outcomes.

\section{Discussion}
\label{S:conclusions}

We have presented a comprehensive representational framework for the operational analysis of neuroradiological reports. Its primary objective is enabling quantitative monitoring sensitive not only to the volume and type of neuroradiological activity but---crucially---its rich content. This functionality provides greater precision and flexibility in the optimisation of service delivery, with a focus on the specifics of the clinical caseload. Our representational approach renders the variety of reported radiological appearances readily survey-able, facilitating operational decision-making responsive to finely nuanced differences in activity.

Deriving actionable representations of radiological reports relies on language modelling but is not achieved by it alone. Our framework makes use of the characteristic structure of reports to identify the assertion and relevant denial of key imaging features as well as their relations: to each other and to the underlying anatomy. This yields a rich representation of the space of possible radiological appearances beyond diagnostic labels (whose derivation imaging in any event naturally precedes) that captures their co-occurrence, both pathological and anatomical. This representation is rendered inspectable by projection to a two-dimensional latent space---employing deep autoencoding to maximise its expressivity---where topological inference can be used to define the dependencies of distinct constellations of appearances on variables of interest. Changes in contrast utilisation, for example, can be interpreted in the context of the reported appearances likely to influence them, not merely the originating clinical pathway. This process allows us to place each report into a learned context, providing a means of interrogating the content and structure of each report relative to all others: a form of data-driven computational phenotyping \cite{Hodapp2016,yang2020combining,kim2020validation} here used for operational purposes but potentially valuable in other contexts. 

Representational analysis is preceded by identification of global features that qualify a report's informativity: assertions of artefact limiting the interpretability of the underlying image, of comparison rendering the report potentially elliptic, and of overall normality eliminating the need for a detailed interpretation. This ensures that each representation captures the radiologist's belief or intention on the applicable limits of description, providing a meta-epistemic index while being of operational interest in its own right. The framework facilitates operational phenotyping by identifying patterns in report content that correlate with clinical practices, such as contrast usage or treatment modalities. For example, significant intersections between neoplastic reports and treatment domains were observed, providing insights into how imaging data informs therapeutic decisions. GeoSPM-based topological inference identified regions of significant association between variables such as patient age and ischaemia or treatment and neoplasia. These findings underscore the utility of spatial inference in linking radiological appearances with demographic or clinical factors. 

Model interpretability is crucial in healthcare to ensure trust and safe decision-making \cite{Yoon581}. This study employs techniques such as 2D latent space visualization, feature attribution, and rule-based systems to enhance transparency. Latent space embeddings reveal patient phenotypes and pathological domains, while feature attribution highlights key factors driving predictions. Rule-based components, like those for negation detection, provide explicit, clinically relevant outputs. For example, spatial inference methods identify regions linked to contrast agent usage, enabling validation against clinical practices. While these approaches align model outputs with clinical intuition, limitations like latent space oversimplification and rule rigidity highlight the need for future work on advanced explainable AI techniques.  

By transforming unstructured text into structured, inspectable representations, it enables a deeper understanding of radiological data beyond traditional diagnostic labels. This has several implications, firstly in clinical decision support. The ability to identify clusters and intersections within the latent space can inform clinicians about common co-occurrences of pathologies or treatment responses, enhancing diagnostic accuracy and therapeutic planning. Secondly, operational efficiency can be helped by the framework's capacity to stratify reports by modality, contrast usage, or clinician ordering patterns provides actionable insights for optimizing radiology workflows and resource utilization. Thirdly, generalizability: the demonstrated performance across multiple institutions highlights the scalability of the framework for broader deployment in diverse healthcare settings. 

This study is limited by the closed nature of the available data, with reliance on data from two institutions potentially restricting generalizability. The dataset size for each sub-task could be expanded to improve performance across different pathological domains. Additionally, the scarcity of domain-specific training data, exacerbated by privacy constraints and the high cost of expert manual labelling, poses challenges. Biases may also arise from reliance on data from a single site, further affecting generalizability to other institutions or reporting styles. Reliable anonymization of clinical data remains difficult, particularly when proper names serve multiple purposes. Moreover, rarer or more heterogeneous pathological domains are less well-represented in the model's latent space, potentially limiting its performance in these areas. Beyond these issues, concerns include the interpretability of the model for clinical users, scalability due to computational demands, and the risk of temporal degradation in performance as clinical practices evolve. Ethical considerations, such as addressing biases in training data and ensuring responsible use of predictive models in healthcare, also warrant attention.

The field of medical natural language processing must account for concerns regarding patient privacy and the automation of medical report analysis. By employing strict anonymization we ensure no identifiable information is fed to the training routine. Furthermore, the pipeline is designed for deployment within the system of a healthcare institution that enables total control of sensitive data. To tackle concerns about model interpretability, the framework integrates rule-based and machine-learning approaches that explicitly model relationships between pathological features, anatomical locations, and diagnostic conclusions. A two-dimensional latent space for report embeddings enhances transparency, allowing clinicians to visualize and validate patterns in the data. By aligning with established ontologies and maintaining human oversight, the system ensures accountability, safeguards patient privacy, and fosters trust in the automation of medical report analysis.

The latent space representations derived from the reports enable the identification of distinct patterns in pathological appearances and their relationships with clinical variables such as patient age, treatment types, and imaging modalities. These representations allow for operational phenotyping, enabling healthcare providers to stratify patient populations based on nuanced imaging features rather than broad diagnostic categories. For example, clusters representing specific conditions (e.g., small vessel disease or neoplasia) can guide targeted interventions or resource allocation.  The system facilitates the integration of radiological findings with patient outcomes (e.g., ICD-10 codes), bridging the gap between imaging data and broader clinical decision-making. Neuradicon exemplifies how AI can enhance radiology by transforming qualitative data into quantitative insights, thereby enabling more precise monitoring and optimization of radiological services. Its ability to produce rich, interpretable phenotypes from text data positions it as a critical tool for advancing personalized medicine and improving healthcare delivery efficiency.

In the broader landscape of medical AI, Neuradicon represents a significant step toward making unstructured clinical data actionable at scale. Its reproducibility across institutions addresses a common barrier in AI adoption—variability in data quality and reporting practices—while its scalability ensures applicability to diverse healthcare settings. By providing a blueprint for integrating NLP with radiology workflows, this framework underscores the potential of AI to transform service delivery in radiology and beyond.

\section{Conclusions}

Recent advances in the field of large language models have demonstrated remarkable improvements in a wide variety of NLP tasks, including in the medical domain \cite{thirunavukarasu2023large}. Models such as GatorTron \cite{yang2022large} and Med-PaLM\cite{singhal2023large} have shown strong performance in medical question-answering tasks and text generation in a medical setting. Trained on general corpora, such large models require computationally expensive tuning to operate in highly specialised domains such as neuroradiology, and do not provide easy mechanisms for the high degree of alignment a clinical setting requires, even in an operational context. 

This study highlights the potential of the Neuradicon framework in converting unstructured neuroradiological reports into structured, actionable insights for clinical and operational use. By employing a hybrid approach that combines rule-based and machine-learning models, the framework demonstrated exceptional performance across tasks such as report classification (F1-score: 0.96), section classification (F1-score: 0.93), and negation detection (F1-score: 0.93). Utilizing autoencoders for two-dimensional latent space representation, Neuradicon identified distinct clusters corresponding to pathological domains (e.g., ischaemic,  haemorrhagic, and neoplastic conditions) and uncovered novel imaging phenotypes based on shared feature patterns. GeoSPM-based spatial inference further revealed significant intersections between variables like age, treatment, and pathology, offering operational insights into resource allocation and planning. The framework’s scalability was validated through its robust generalizability across two healthcare institutions (UCLH and KCH), achieving consistent high performance on prospective data. Additionally, its ability to stratify reports by modality, contrast usage, or clinician ordering patterns bridges clinical decision-making with workflow optimization, underscoring its value in enhancing both radiological practice and healthcare delivery. 

Our approach of employing an inspectable representation promotes trust by rendering the organisation of reports relative to one another recognisable to clinicians \cite{nix_understanding_2022}. Indeed, it can facilitate the validation of image analytic models that output machine-generated reports by enabling their comparison with expert generated reports in the latent space. To further enhance its utility and adoption, future work should focus on addressing limitations such as variability in reporting styles across institutions and improving interpretability for non-technical users. Additionally, integrating multimodal data (e.g., imaging features) with textual analysis could provide even richer insights into patient care pathways. The Neuradicon framework offers a blueprint for using natural language processing (NLP) to unlock the value of unstructured medical text data in neuroradiology. Its ability to extract quantitative insights from qualitative data has implications for: 

\begin{itemize}
    \item \textbf{Personalized Medicine}: By identifying patient-specific phenotypes and patterns of disease progression, Neuradicon can support tailored therapeutic strategies.
    \item \textbf{Operational Efficiency}: Insights into workflow patterns can inform resource allocation, scheduling optimization, and quality assurance in radiology departments.
    \item \textbf{Scalability to Other Domains}: While this study focused on neuroradiology, the methods are generalizable to other medical fields where unstructured text data is prevalent.
\end{itemize}
  
Operational objectives aside, Neuradicon is in theory applicable to the tasks of organising historical radiological corpora for automated analysis, triggering automated image-analytic routines conditional on the presence of specific reported features, enabling multimodal modelling of the imaged brain, and identifying characteristic descriptive patterns of research and educational interest.

\backmatter

\section{Declarations}

\subsection{Ethics}
This project was conducted at University College London Hospitals NHS Trust as a service evaluation and optimization project employing irrevocably anonymised data, which does not require research ethics approval since its primary objective is not research but service improvement. KCH data was sourced from the LMIAI Centre for Value-Based Healthcare Anonymised Database, operating under ethical approval from East of Scotland Research Ethics Service (EoSRES). REC reference: 20/ES/0005 IRAS project ID: 257568.

\subsection{Data Availability}
In keeping with internal service projects, the terms of access to source data preclude their public dissemination. 

\subsection{Code Availability}
The software used in this work were as follows: Python 3.8 (https://www.python.org/), scikit-learn 0.23 (https://scikit-learn.org), Numpy 1.21 (https://numpy.org), Scipy 1.7 (https://www.scipy.org/), Pandas 1.3.1 (https://pandas.pydata.org/), Spacy 3.0 (https://spacy.io/), prodigy 1.0 (https://prodi.gy/), Huggingface (https://huggingface.co/), geoSPM (https://github.com/high-dimensional/geospm). 
All software are open source and publicly available.

\subsection{Author Contributions}
H. W., R. G. and P. N. contributed to the design of the study. H. W. is responsible for the creation of the software and models. A. Jha, Y. M., R. G., H. W., W.H.L.P., J. T contributed to the acquisition and analysis of the data, including labelling and evaluation. H.W., A. Julius, and P.N. prepared the manuscript, which was reviewed and approved by all authors. 

\subsection{Competing Interests Statement}
The authors declare no competing interests

\subsection{Acknowledgements}
This work was funded by the Wellcome Trust via an Innovations Project Award Ref. 213038/Z/18/Z. Y.M. is funded by an MRC grant (MR/T005351/1). Additionally this project was supported by the UCLH NIHR biomedical research centre. 

\bibliography{used_references}

\appendix

\section{Data Collection}
\label{app:data}

To address potential confounding factors and biases, such as variations in radiologist reporting styles and image acquisition methods, this study implemented several robust strategies to ensure the reliability and generalizability of its findings. By incorporating cross-site data from two institutions (UCLH and KCH), the models were designed to avoid overfitting to site-specific characteristics, thereby enhancing their applicability across diverse clinical settings. The use of rule-based methods like Negbio for negation detection further mitigated biases by leveraging grammatical patterns rather than relying solely on potentially skewed training data. To ensure consistent labelling and high inter-rater reliability, a consensus-annotation process was employed across datasets. Additionally, spatial inference techniques, such as GeoSPM (see section \ref{S:geospm}), were applied to account for confounding variables like patient age and treatment domains, enabling precise statistical mapping of significant patterns. Finally, a latent space embedding (see section \ref{S:embedding}) was utilized to cluster similar pathological features while minimizing the influence of stylistic differences in reporting. Together, these comprehensive measures effectively addressed biases and ensured robust performance across variations in style, authorship, and institutional practices.

Labelled data for each task was produced using the prodigy labelling tool (https://prodi.gy/).  Each report was labelled in a consensus-annotation manner, whereby two annotators pass through each item at the same time and label a report or instance once consensus has been reached. This consensus labelling method ensures inter-observer variability is minimised at source. For the negation labels, each entity in a report was labelled as being either asserted or denied, thus each entity had a binary label. For the negation detection and relation extraction tasks, the classification is performed at the level of individual entities. The total number of labelled instances is much larger than the number of reports in the test set. The relation extraction labels are acquired by using the relation labelling mode in the prodigy software, where links between different spans are identified in the text. Furthermore, the negation detection and relation extraction models are rules-based so only an evaluation split was required.

Data for the report classification task was produced by labelling each report with an exclusive class. The total dataset was then split into train/evaluation sets with a 80/20 split. The section segmentation data is labelled using the initial token of the section sequence. Figure \ref{fig:section} shows a schematic of how the label scheme works. The token corresponding to the start of each section is labelled with the section type with all other tokens unlabelled. As a token classification model, the number of individual samples in the section classification dataset is much larger than the number of reports. This data was split into training/evaluation sets with 80\% of reports in the train set and 20\% of reports in the evaluation set.

The domain classification datasets were acquired from report corpora from both UCLH and KCL. Each is further subdivided into a 'current' dataset and a 'prospective' dataset. There are thus 4 test datasets for the domain classification task. This split is to ensure robust performance across time as well as across variation in style and content. Each report is labelled in a non-exclusive, multi-label manner with all the domain types relevant for that report. This task does not require a training dataset because predictions are derived from the output of the NER and negation detection models.

\subsection{Example Report}
\label{app:report}
"There is moderate dilatation of the third and lateral ventricles, distension of the third ventricular recesses and mild enlargement of the pituitary fossa with depression of the glandular tissue. Appearances are in keeping with hydrocephalus and the mild associated frontal and peritrigonal transependymal oedema indicates an ongoing active element. No cause for hydrocephalus is demonstrated and the fourth ventricle and aqueduct are normal in appearance. Note is made of T1 shortening within the floor of the third ventricle. On the T2 gradient echo there is a bulbous focus of reduced signal in this region that is inseparable from the terminal basilar artery but likely to be lying anterosuperiorly. The combination of features is most likely to reflect a small incidental dermoid within the third ventricle with fat and calcified components. This could be confirmed by examination of the CT mentioned on the request form. If the presence of fat and calcium is not corroborated on the plain CT, a time-of-flight MR sequence or CTA would be prudent to exclude the small possibility of a vascular abnormality in this region."

\section{Model training details}

\subsection{Language model}
\label{app:llm}
The neuradicon pipeline makes use of the huggingface library for training a custom BioBERT LLM. This is trained unsupervised by fine-tuning the model according to the suggested the parameters defined in the huggingface documentation (https://huggingface.co/docs). We used the masked-language-modelling object for 100 epochs with a batch size of 32 and a learning rate of $0.0002$. The model was trained in float-32 precision using the Adam optimizer. The choice of precision, optimizer and learning rate are the setting recommended by the huggingface library. Some trail and error was performed regarding the batch size and learning rate, but the overall performance during tuning was unaffected by a particular batch size.
This base language model produces 512-dimensional token vectors that are used as the input to downstream models. 
\subsection{Text classifier model}
\label{app:cls}
The report classification model is a text classification model implemented using the text categorisation tool of Huggingface \cite{jain2022hugging}. The approach uses the AutoModelForSequenceClassification class with a 5-class categorical cross-entropy loss. The optimizer implements the Adam optimiser \cite{Kingma2015adam}, with a learning rate of $0.001$ and a batch size of 32. There is a period of warmup of 1000 steps where the learning rate gradually grows linearly to the learning rate of $0.0001$. It also implements a weight decay of $0.001$ to impose regularization during traing. Furthermore, the model was also trained with dropout with a value of $0.4$ to improve generalization.
\subsection{Section segmentation model}
\label{app:cls2}
The section segmentation task is implemented using the token classification model from the Huggingface library. Using the base language model of section \ref{S:word2vec}, the token classification layer takes the token representations and classifies them into correct classes. We take a AutoModelForTokenClassification model and try to predict those tokens that correpsond to the start tokens of the corresponding sections. In practice this model correspoinds to a simple feed-forward classification layer on top of the last transformer layer output token activations, passed through a cross-entropy loss function. We train the model using batch sizes of 32, a learning rate of $0.0001$ with Adam optimisation. As with the classification model, we also implement weight decay of $0.001$ and a warmup of 1000 steps and use a dropout value of $0.4$.

\subsection{NER model}
\label{app:ner}
The named entity recognition model in the pipeline implements the AutoModelForTokenClassification from the HuggingFace library. The training parameters are the same as the above segmentation model parameters, however the data is labelled according to the BIO schema.

\subsection{The Autoencoder model}
\label{app:ae}
The input to the autoencoder model is a high-dimensional binary vector. For an  $N$ dimensional binary input $x \in {0,1}^N$, the autoencoder consists of an encoder $g$, that maps $x$ to a 2 dimensional latent vector $z \in R^2$ and a decoder $f$ that attempts to reconstruct ($x'$) the original input vector.
\begin{align}
z = g(x)\\
x'= f(z)
\end{align}
Our autoencoder is implemented using Pytorch. The autoencoder is composed of 5 hidden layers with sizes (256,64,2,64,256) separated by batch normalisation \cite{batchnorm} ($bn_i$) and ELU activation \cite{ELUclevert}, with a final output sigmoid to ensure the output is within [0,1]. Thus the encoder $f$ is made up of linear layers $f_i$,
\begin{align}
    f(x) = f_3(ELU(bn_2(f_2(ELU(bn_1(f_1(x))))))),
\end{align}
and the decoder the mirror image of $f$. The model is trained with a mean squared error loss function and Adam optimisation with a learning rate of $10^-4$. We train this model without any weight decay or dropout until convergence.
\end{document}